\documentclass{article}
\usepackage{amsmath}
\usepackage{booktabs}
\usepackage{array}
\usepackage{caption}
\usepackage{graphicx}
\usepackage{xspace}
\usepackage{pifont}
\newcommand{\cmark}{\ding{51}}
\newcommand{\xmark}{\ding{55}}
\usepackage{xcolor}
\usepackage[colorlinks=true, citecolor=blue, linkcolor=blue, urlcolor=blue]{hyperref}
\usepackage{wrapfig}
\usepackage{enumitem}
\usepackage{multirow}

\usepackage[most]{tcolorbox}
\usepackage{listings}

\lstdefinestyle{promptstyle}{
    basicstyle=\ttfamily\footnotesize,
    breaklines=true,
    breakatwhitespace=true,
    columns=fullflexible,
    keepspaces=true,
    showstringspaces=false,
    frame=none
}

\usepackage[preprint]{corl_2026} % Uncomment for pre-prints (e.g., arxiv); This is like ``final'', but will remove the CORL footnote.

\definecolor{citecolor}{HTML}{0071BC}
\hypersetup{colorlinks,linkcolor={red},citecolor={citecolor}}  

\newcommand{\ours}{DenseReward\xspace}

\title{\ours: Dense Reward Learning via Failure Synthesis for Robotic Manipulation}

\author{
    Yu Fang$^{1}${\hypersetup{linkcolor=black}\thanks{Equal contribution.}} \quad
    Wanxi Dong$^{1}$\footnotemark[1] \quad
    Jiaqi Liu$^{1}$ \quad
    Yue Yang$^{1}$ \quad
    Mingxiao Huo$^{2}$ \\ 
    \textbf{
        Yao Mu$^{3}$ \quad
        Huaxiu Yao$^{1}$ \quad
        Li Erran Li$^{4}$ \quad
        Daniel Szafir$^{1}$ \quad
        Mingyu Ding$^{1}$
    }
    \\
    $^{1}$University of North Carolina at Chapel Hill \quad
    $^{2}$Carnegie Mellon University
    \\
    $^{3}$Shanghai Jiao Tong University \quad
    $^{4}$Amazon AWS AI
    \\
    {\href{https://dense-reward.github.io/}{\textcolor{blue}{https://dense-reward.github.io/}}}
}

\begin{document}
\maketitle

%===============================================================================

\begin{figure}[h]
    \centering
    \vspace{-24pt}
    \includegraphics[width=1.0\linewidth]{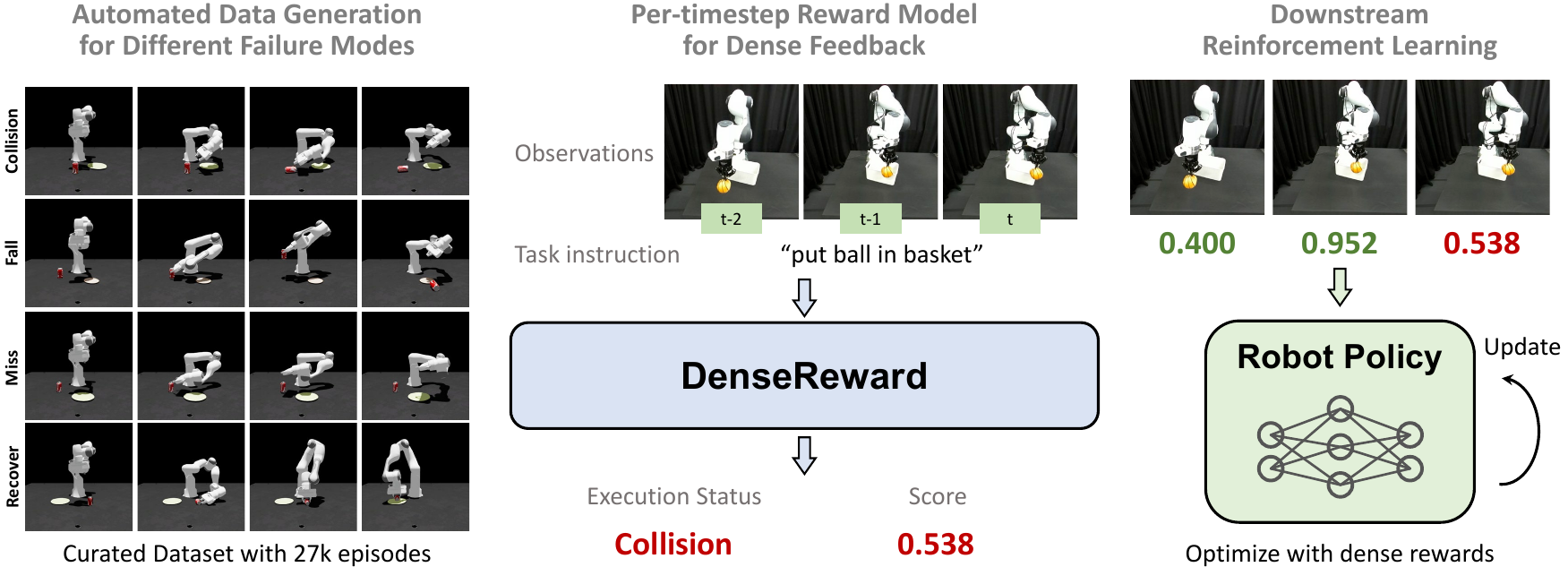}
    \vspace{-8pt}
    \captionof{figure}{\textbf{Overview.}
        We present a dense robotic reward model for vision-language-guided manipulation.
        \textbf{(a)} We construct a 27k-episode dataset with automatically synthesized success and failure trajectories in simulation, providing diverse dense reward supervision.
        \textbf{(b)} Given a task instruction and observations consisting of the current frame and historical frames, \ours predicts a per-timestep dense reward score that reflects task progress.
        Unlike binary success/failure labels, this dense reward provides fine-grained feedback for intermediate states.
        \textbf{(c)} The predicted rewards can be used in downstream reinforcement learning for policy improvement.}
    \label{fig:teaser}
\end{figure}
\vspace{-8pt}

\begin{abstract}
    Reinforcement learning holds great promise for improving robot policies beyond the limits of imitation learning.
    However, its practical adoption remains bottlenecked by the lack of reliable vision-language reward models that provide dense and informative feedback.
    Two key challenges remain: acquiring diverse failure data at scale and obtaining fine-grained reward signals beyond sparse trajectory-level success labels.
    Existing approaches reuse successful trajectories to construct negative or suboptimal examples through trajectory truncation, instruction relabeling, or trajectory comparisons, without generating new trajectories where physical interaction failures such as collisions, missed grasps, object drops, and recovery actually occur.
    Meanwhile, many existing reward models predict sparse binary or trajectory-level rewards, which provide limited guidance for efficient policy optimization.
    We introduce \ours, a dense robotic reward model that addresses both challenges.
    To train \ours, we develop an automated failure data generation pipeline that synthesizes physically realistic failure trajectories in simulation without manual annotation, covering diverse failure modes such as collisions, missed grasps, object drops, and recovery behaviors.
    \ours predicts dense frame-level reward scores from visual observations and language instructions, enabling fine-grained estimation of task progress throughout an episode.
    Experiments show that \ours outperforms general-purpose VLMs and existing robotic reward models in dense reward prediction across both simulated and real-world manipulation.
    We further demonstrate that \ours provides effective reward guidance for downstream model predictive control and reinforcement learning.
    We release the dataset, trained reward models, and evaluation suite to support the development of failure-aware dense reward modeling for robot learning.
\end{abstract}


%===============================================================================

\section{Introduction}
Robotic manipulation has seen remarkable progress through imitation learning on large-scale demonstration datasets~\cite{brohan2022rt, walke2023bridgedata, o2024open, khazatsky2024droid}.
Reinforcement learning (RL) offers a complementary path beyond imitation, enabling robot policies to improve through trial-and-error interaction with the environment.
Recent works have applied on-policy RL algorithms~\cite{schulman2017proximal, shao2024deepseekmath} to finetune vision-language-action models~\cite{black2024pi_0, kim24openvla}, showing that RL can push policies beyond the performance ceiling imposed by demonstration data~\cite{wagenmaker2025steering, intelligence2025pi06}.
However, current approaches often rely on sparse binary rewards provided only at episode end, which suffer from severe credit assignment problems in manipulation, making policy learning sample inefficient and difficult to scale.
Dense reward signals that provide per-timestep feedback on task progress offer a more informative learning signal, but designing such rewards remains an open and largely unsolved problem.

Recently, Vision-language models (VLMs) have emerged as a promising foundation for general-purpose robotic reward models~\cite{mahmoudieh2022zero, sontakke2023roboclip, wang2024rl, zhao2024vlmpc, ma2025gvl, tan2025robo, zhai2025vision, venkataraman2025real, chen2026topreward, liang2026robometer}, by leveraging broad semantic knowledge from large-scale pretraining.
Despite this promise, existing vision-language reward models face two fundamental limitations. 
First, reward models require diverse failure trajectories to learn failure modes and the reasons that lead to failures, yet they are primarily trained on large-scale datasets that only include successful demonstrations~\cite{o2024open, lee2026roboreward}.
Recent works address this issue through data augmentation: RoboReward~\cite{lee2026roboreward} truncates successful episodes at intermediate frames, while Robometer~\cite{liang2026robometer} constructs preference pairs from suboptimal rollouts.
These approaches provide scalable negative or suboptimal supervision derived from existing trajectories.
However, they do not explicitly capture physical interaction events, such as missed grasps, collisions, object drops, or recovery behaviors.
A complementary direction is therefore to directly generate trajectories in which such events occur during execution.
Second, many existing reward models produce sparse trajectory-level reward signals, assigning a single label to an entire rollout at the end~\cite{ lee2026roboreward, sontakke2023roboclip}.
The policy receives no feedback about which intermediate actions contributed to success or failure, making it difficult to guide policy optimization over long-horizon manipulation tasks.

In this paper, we introduce \ours, a dense vision-language reward model for robotic manipulation.
We address two challenges: acquiring diverse failure data at scale and providing dense reward signals for downstream policy learning.
We develop an automated data generation pipeline in Isaac Sim that produces physically realistic successful and failure trajectories without manual trajectory annotation.
We structure manipulation into five canonical phases: \emph{Reach}, \emph{Grasp}, \emph{Lift}, \emph{Move}, and \emph{Place}, and apply targeted perturbations at different execution stages to induce interactive events such as collisions, missed grasps, object drops, suboptimal motions, and recovery behaviors.
In contrast to truncating or relabeling successful demonstrations, these simulator-generated trajectories explicitly contain the corresponding physical interaction events and their temporal consequences.
We assign per-timestep reward scores that represent task progress throughout execution, including partial completion, degradation after failure events, and recovery from temporary failures.
We combine the Isaac Sim trajectories with additional trajectories from DROID, RoboSuite, and LIBERO to construct a dataset of 27k episodes spanning diverse objects, tasks, and visual environments.
Given a language instruction, the current observation and historical frames, \ours predicts the current execution status (success or a specific failure type) and a scalar score that estimates task progress at the current timestep.
Our contributions are threefold.
\begin{itemize}[leftmargin=*, noitemsep, topsep=0pt]
\item \textbf{Automated data generation with dense rewards.} We decompose manipulation into five canonical phases and synthesize physically realistic success and failure trajectories across six labeled execution categories through targeted perturbations of grasp detection and motion planning, producing dense per-timestep reward labels and labeled failure reasons without manual annotation.
\item \textbf{\ours dataset and model.} We curate a dataset of 27k episodes spanning both successful and failed trajectories across diverse tasks and scenes. On top of this, we train \ours, a vision-language reward model that predicts both the current execution state and task progress.
\item \textbf{Downstream applications.} \ours achieves lower dense reward prediction error than general-purpose VLMs and existing robotic reward models on our evaluation set. We further demonstrate that \ours provides effective reward guidance for local control and serves as a shaping signal for reinforcement learning in both simulation and the real world.
\end{itemize}

\section{Related Work}
\label{sec:relatedwork}

\textbf{Reward Models for Robotic Manipulation.}
Designing reward functions for robot manipulation has been a central challenge~\cite{sadigh2017active, mahmoudieh2022zero}.
Classical approaches rely on hand-crafted reward shaping~\cite{kumar2016optimal, rajeswaran2018dexterous}, which requires task-specific engineering and can be difficult to transfer across tasks.
VLM-based reward models leverage broad semantic knowledge from pretraining to provide task-conditioned evaluation without manual engineering~\cite{ma2022vip, sontakke2023roboclip, ma2023liv, wang2024rl, duan2024aha,rocamonde2024vision,ma2025gvl, tan2025robo, zhai2025vision, chen2026topreward, zhang2025reinbot, chen2025sarm,liang2026robometer,schroeder2026sole}.
Many existing reward models focus on sparse trajectory-level signals~\cite{sontakke2023roboclip, chen2026topreward, lee2026roboreward}, while recent work additionally explores denser supervision through preference data, progress annotations, and suboptimal rollouts~\cite{liang2026robometer}.
\ours is complementary to these approaches: rather than constructing supervision only from existing successful trajectories, we synthesize trajectories with explicit interaction events such as collisions, missed grasps, object drops, and recovery, and use them to train a model for per-timestep progress prediction.

\textbf{Failure Data Generation.}
Failure data is valuable for learning robust robot policies and reward models, but collecting diverse failures at scale remains challenging.
Human teleoperation can provide realistic failure trajectories~\cite{mandlekar2019scaling, ye2025robofac, wu2025robomind}, but requires substantial human effort.
Existing data augmentation approaches construct negative or suboptimal examples from successful demonstrations~\cite{lee2026roboreward, glossop2025cast, zhang2025rewind, Dai2024racer}, but do not generate executions in which physical failure events occur.
Simulation offers a complementary mechanism for directly inducing specific interaction events under controlled conditions
~\cite{tobin2017domain, mandlekar2017adversarially, pacaud2024failcot}.
We apply targeted perturbations during simulated execution to produce failure trajectories with dense reward annotations and execution status labels.

\textbf{Reinforcement Learning for VLAs.}
Reinforcement learning improves robot policies beyond imitation learning~\cite{haarnoja2018soft, wagenmaker2025steering,intelligence2025pi06, zhang2025prophrl, li2025simplevla}.
Recent works apply on-policy algorithms~\cite{schulman2017proximal, li2025adaptive,guo2025irevla,intelligence2026pi07} to finetune pretrained VLAs and achieve promising gains~\cite{black2024pi_0, kim24openvla,agibot2026sop,huo2024sparse, li2025gr,chen2025rlrc,wang2026reinforcing}.
DSRL~\cite{wagenmaker2025steering} steers the diffusion noise space of pretrained policies, enabling stable real-world RL adaptation under limited rollout budgets.
However, these approaches typically rely on sparse binary rewards at episode termination, which provide limited credit assignment for intermediate actions.
World models and video generation models predict future observations to provide dense reward signals~\cite{ebert2018visual,hafner2019dream,escontrela2023viper, zhang2025prophrl,guo2025ctrl,peng2026reworld, he2026pre, jiang2026wovr, lv2026viva}, but can be unreliable in contact-rich manipulation.
\ours predicts dense rewards directly from visual observations and language instructions, enabling effective RL finetuning on both simulated and real-world tasks.

%===============================================================================

\begin{figure}
    \centering\includegraphics[width=0.95\linewidth]{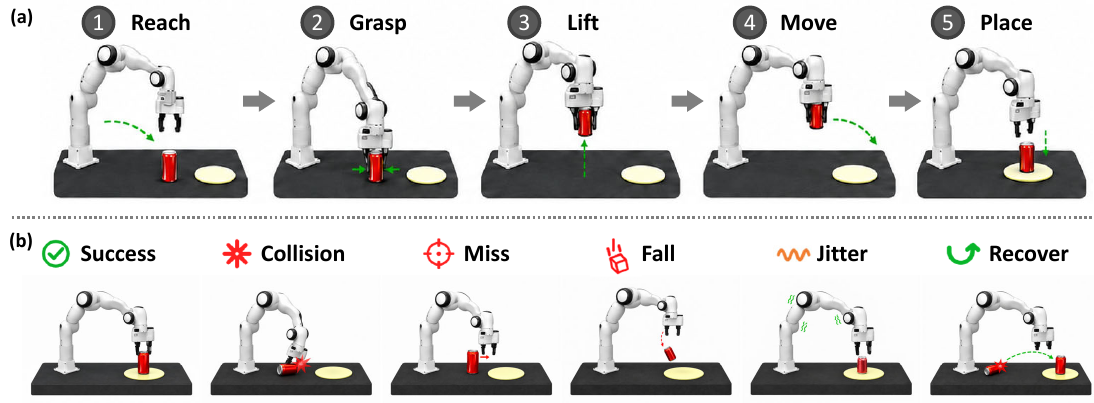}
    \makebox[\linewidth][l]{%
        \hspace*{0.025\linewidth}%
        \includegraphics[width=0.95\linewidth]{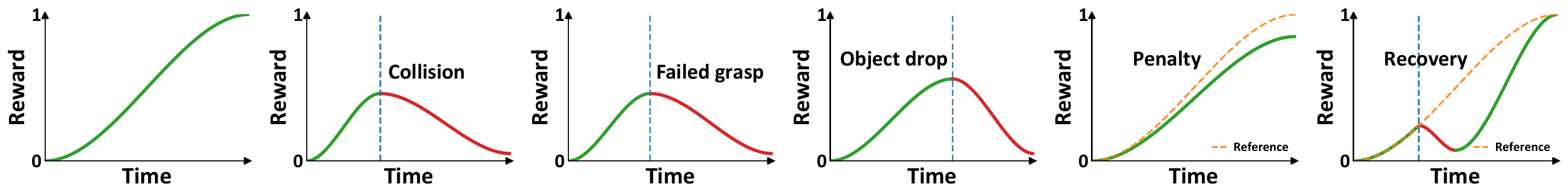}%
    }
    \vspace{-8pt}
    \caption{\textbf{Overview of Automated Data Generation and Failure Synthesis.}
    (a) Five-phase manipulation with dense rewards.
    (b) Targeted perturbations synthesize diverse failure modes.
    }
    \label{fig:method}
    \vspace{-8pt}
\end{figure}

\section{Method}
\label{sec:method}
We present \ours, a vision-language reward model that predicts dense rewards for robotic manipulation.
Our method consists of three components:
1) an automated data generation pipeline that generates trajectories with phase-aware dense reward labels,
2) failure synthesis that creates diverse failure trajectories through targeted perturbations, and
3) \ours models trained on the resulting mixture of successful and failure trajectories to estimate fine-grained task progress.

\subsection{Automated Data Generation}

\textbf{Dense Reward Formulation.}
Instead of assigning a single label $p \in [0, 1]$ at the end of an entire trajectory, we aim to learn reward signals at every timestep throughout execution.
For each trajectory $\tau = \{l, \mathbf{o}_{1:T}, \mathbf{r}_{1:T}\}$ that contains a language instruction $l$ and image observations $\mathbf{o}_{1:T}$, we define per-timestep dense rewards $\mathbf{r}_{1:T}$ with $r_t \in [0, 1]$. 

\textbf{Phase Decomposition.}
To construct $\mathbf{r}_{1:T}$ without manual annotation, we decompose manipulation into five canonical phases:
1) \textbf{Reach}: the robot moves its end-effector toward the target object;
2) \textbf{Grasp}: the robot closes its gripper to secure the object;
3) \textbf{Lift}: the robot raises the object off the surface;
4) \textbf{Move}: the robot transports the object toward the target location;
5) \textbf{Place}: the robot releases the object at the goal pose.
This provides a structured basis for automatically generating trajectories with dense rewards.

\textbf{Automated Pipeline.}
Based on these phases, we design an automated pipeline for trajectory generation, as shown in Fig.~\ref{fig:method}(a).
For each episode, we randomly initialize the scene by placing the target object and goal container at different positions on the table, producing diverse spatial configurations across episodes.
We then use GraspNet~\cite{fang2020graspnet} to predict up to $N=50$ candidate grasp poses from multi-view RGB-D observations captured in simulation.
These candidates are passed to CuRobo~\cite{sundaralingam2023curobo} for collision-aware motion planning to select a feasible grasp.
CuRobo then plans the motions to execute the five manipulation phases, which the robot follows sequentially.
Phase boundaries are automatically determined from simulation state: Grasp begins when the gripper contacts the object, Lift begins once the object leaves the table, and Place begins when the end-effector comes within a distance $d_{\text{place}}$ of the target location.
The entire process requires no manual annotation.

\subsection{Failure Synthesis}
Building on the automated data generation pipeline, we synthesize failure trajectories with dense rewards, as illustrated in Fig.~\ref{fig:method}(b).

\textbf{Success and Failure Types.}
We first define failure modes and induce failure trajectories by targeted perturbations at specific stages of this pipeline.
1) \textbf{Success.} A complete, unperturbed execution where the robot successfully places the object and the reward rises monotonically.
2) \textbf{Collision.} The robot collides with the environment or object. The reward follows a mountain-shaped curve, reached at the collision event, then decaying.
3) \textbf{Miss.} The gripper fails to grasp the object. Similarly to Collision, the reward rises until the failed grasp attempt, then decays.
4) \textbf{Fall.} The robot successfully grasps and lifts the object, but the object falls before reaching the target. The reward follows a mountain-shaped curve, rising to a peak at the drop event then decaying, reflecting partial progress that is ultimately unsuccessful.
5) \textbf{Jitter.} An execution with penalized scaled reward, representing trajectories with suboptimal motion. This captures that task completion via an inefficient path is penalized compared to a precise execution.
6) \textbf{Recover.} The robot encounters a collision but successfully recovers and completes the task. The reward drops while the collision happens, then resumes climbing once the robot clears the obstruction, capturing the temporal penalty of recovery.

\textbf{Perturbations.}
We design perturbations to induce failure trajectories for each mode at specific stages.
1) \textbf{Success.} No perturbations.
2) \textbf{Collision.} The motion is planned with collision avoidance disabled, forcing the robot through an infeasible path that hits the object or table.
3) \textbf{Miss.} We offset the grasp target pose, which causes the gripper to close in the air, preventing a stable grasp.
4) \textbf{Fall.} Random rotation perturbations are applied to the movement during the Move Phase, causing the gripped object to lose stability and fall during transportation.
5) \textbf{Jitter.} We inject a small Gaussian joint noise at every timestep to produce a jittery trajectory.
6) \textbf{Recover.} The robot first encounters a collision, then the motion planner replans a clear path to complete the task.
To filter physically invalid episodes, we apply automated validity checks at the stage boundaries to ensure the object position is reasonable during each phase.
Trajectories failing these checks are discarded.

\subsection{\ours Training}
\textbf{Dataset.} 
We construct a dense reward dataset containing 27k successful and failed manipulation episodes.
Our automated failure synthesis pipeline is applied in Isaac Sim, where we generate 12k trajectories spanning diverse execution outcomes with more than 60 objects.
To increase the diversity of tasks, scenes, and observations, we additionally incorporate trajectories from DROID~\cite{khazatsky2024droid}, RoboSuite~\cite{zhu2020robosuite}, and LIBERO~\cite{liu2023libero}.
The resulting dataset spans both simulated and real-world manipulation settings and is divided into training and test sets for model training and evaluation.

\textbf{\ours Models.}
We build \ours on top of Qwen3-VL-4B-Instruct~\cite{Qwen3-VL} and finetune it to predict dense progress scores from observations and language instructions.
As illustrated in Fig.~\ref{fig:teaser}, given a task instruction, the current observation, and historical frames, \ours predicts both the execution status and a scalar progress score for the current timestep.
Unlike reward models trained primarily with trajectory-level binary success labels, \ours uses dense per-timestep supervision to model intermediate task progress throughout an execution.
Reward targets are $[0, 1]$ and represented with three decimal places to preserve fine-grained variation across timesteps, particularly for long trajectories.
We use these predictions as dense feedback for downstream control and policy optimization.

%===============================================================================

\begin{figure}[t!]
    \centering
    \includegraphics[width=1.0\linewidth]{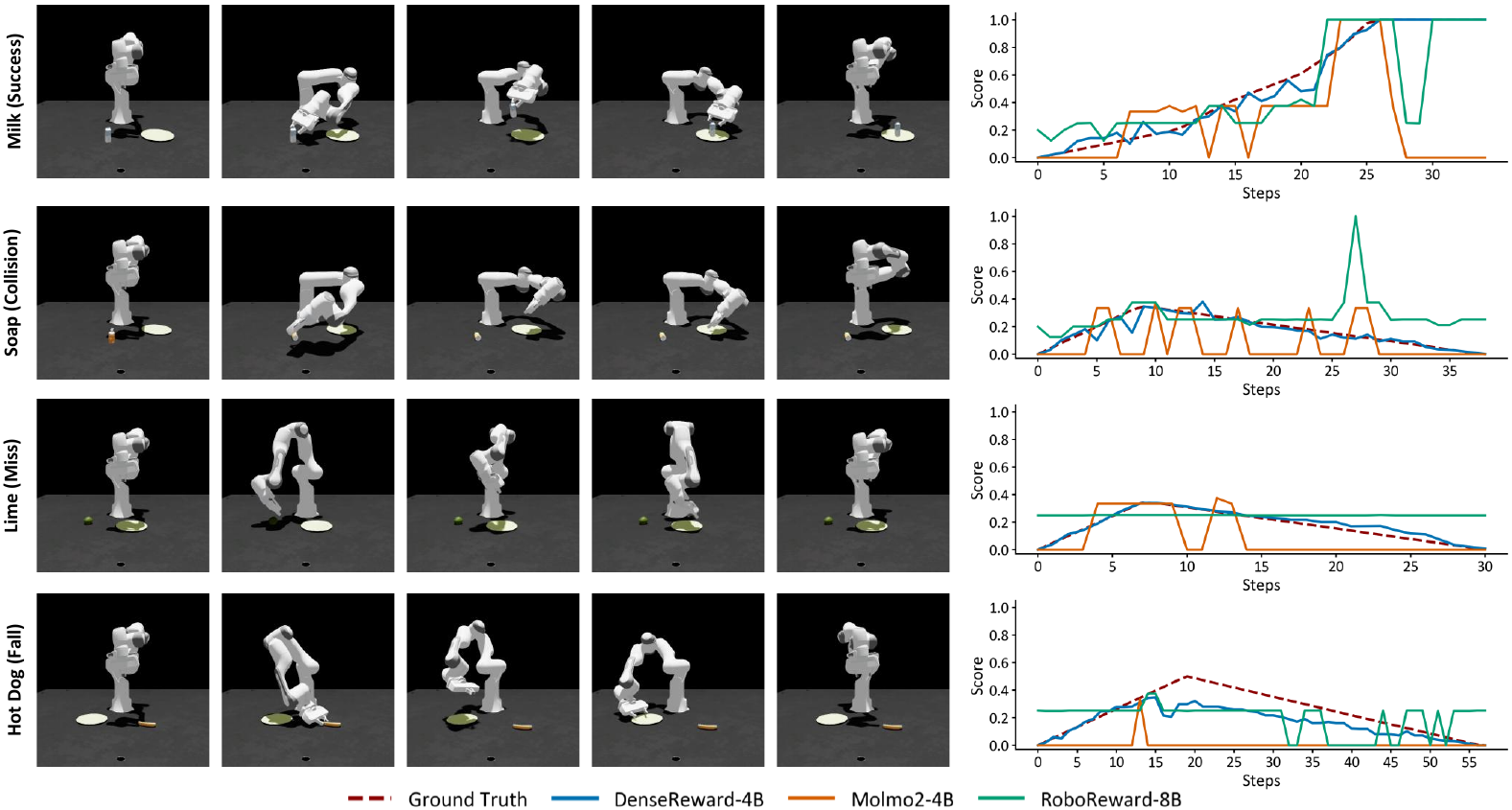}
    \vspace{-12pt}
    \caption{\textbf{Qualitative comparison of dense reward prediction.} \ours more closely follows task progress across successful and failed executions.
    }
    \label{fig:results}
    \vspace{-18pt}
\end{figure}

\begin{table}[t]
\centering
\small
\setlength{\tabcolsep}{6pt}
\renewcommand{\arraystretch}{1.1}
\caption{
    \textbf{Dense reward prediction results.}
    We report the mean absolute error (MAE) with respect to the dense reward annotations in the evaluation set.
    \ours achieves the lowest error across all evaluated data sources.
}
\vspace{0.4em}
\begin{tabular}{>{\centering\arraybackslash}p{3.8cm}|>{\centering\arraybackslash}p{1.2cm}|cccc}
\toprule
\textbf{Model} 
& \textbf{Overall} 
& \textbf{DROID} 
& \textbf{Isaac Sim} 
& \textbf{RoboSuite} 
& \textbf{LIBERO} \\
\midrule
Qwen3-VL-4B-Instruct~\cite{Qwen3-VL} & 0.289 & 0.532 & 0.285 & 0.195 & 0.478 \\
Qwen3-VL-8B-Instruct~\cite{Qwen3-VL} & 0.293 & 0.538 & 0.305 & 0.180 & 0.502 \\
Molmo2-4B~\cite{clark2026molmo2}            & 0.282 & 0.506 & 0.282 & 0.187 & 0.478 \\
Molmo2-8B~\cite{clark2026molmo2}            & 0.335 & 0.480 & 0.307 & 0.303 & 0.455 \\
RoboReward-4B~\cite{lee2026roboreward}        & 0.275 & 0.534 & 0.269 & 0.179 & 0.470 \\
RoboReward-8B~\cite{lee2026roboreward}        & 0.230 & 0.484 & 0.185 & 0.172 & 0.431 \\
Robometer~\cite{liang2026robometer}         & 0.366 & 0.521 & 0.328 & 0.345 & 0.468 \\
\ours (Ours)      & \textbf{0.081} & \textbf{0.259} & \textbf{0.081} & \textbf{0.051} & \textbf{0.044} \\
\bottomrule
\end{tabular}
\label{tab:results}
\vspace{-8pt}
\end{table}

\section{Experiments}
\label{sec:result}
\subsection{Evaluating Reward Prediction}
We evaluate the dense reward prediction performance of \ours against general-purpose vision-language models and existing robotic reward models.
In Tab.~\ref{tab:results}, \ours achieves the lowest overall mean absolute error (MAE) of 0.081 across all evaluated data sources. 
In comparison, Qwen3-VL-4B-Instruct and Qwen3-VL-8B-Instruct obtain overall MAEs of 0.289 and 0.293, respectively, while existing robotic reward models such as RoboReward exhibit larger errors.
Fig.~\ref{fig:results} provides a qualitative comparison with Molmo2-4B and RoboReward-8B, the strongest baselines in their respective categories.
\ours aligns more closely with the task progress, while the baseline predictions are often noisier or less sensitive to fine-grained progress changes.
Together, these results show that \ours more accurately captures the dense progress supervision across diverse manipulation trajectories.

\subsection{Evaluation in Model Predictive Control}

\begin{figure}[t]
    \centering
    \begin{minipage}[t]{0.47\linewidth}
        \vspace{0pt}
        \centering
        \includegraphics[width=\linewidth]{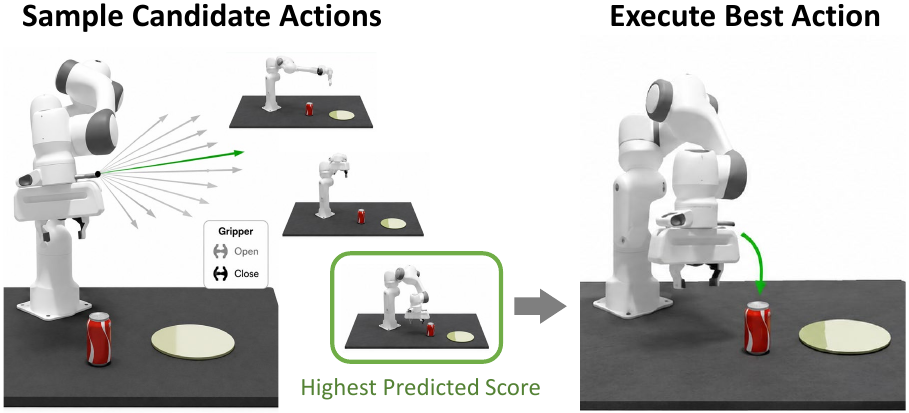}
        \vspace{-10pt}
        \captionof{figure}{\textbf{Dense-reward-guided MPC.}}
        \label{fig:mpc}
    \end{minipage}
    \hfill
    \begin{minipage}[t]{0.50\linewidth}
        \vspace{4pt}
        \centering
        \scriptsize
        \captionof{table}{\textbf{Performance of MPC on three object manipulation tasks.}}
        \label{tab:mpc_results}
        \vspace{2pt}
        \setlength{\tabcolsep}{2.5pt}
        \renewcommand{\arraystretch}{0.95}
        \resizebox{\linewidth}{!}{
            \begin{tabular}{>{\centering\arraybackslash}p{2.6cm}cccc}
            \toprule
            \textbf{Model} & \textbf{Can} & \textbf{Cup} & \textbf{Lemon} & \textbf{Avg.} \\
            \midrule
            RoboReward-4B~\cite{lee2026roboreward} & 0.199 & 0.307 & 0.295 & 0.267 \\
            RoboReward-8B~\cite{lee2026roboreward} & 0.314 & 0.270 & 0.317 & 0.300 \\
            VLAC-2B~\cite{zhai2025vision}          & 0.316 & 0.346 & 0.380 & 0.347 \\
            VLAC-8B~\cite{zhai2025vision}          & 0.351 & 0.360 & 0.363 & 0.358 \\
            \ours (Ours)                            & 0.219 & 0.181 & 0.288 & 0.229 \\
            \bottomrule
            \end{tabular}
        }
    \end{minipage}
    \vspace{-12pt}
\end{figure}

\textbf{Setup.}
We evaluate whether the learned reward signal can provide useful local guidance for action selection in a simple short-horizon control setting.
As shown in Fig.~\ref{fig:mpc}, at each decision step, we sample 28 candidate actions (27 directions and gripper open/close), and use the reward model to score candidate transitions.
The simulator state is then restored, and the action with the highest predicted score is executed.
We evaluate this procedure on three object manipulation settings involving a can, cup, and lemon.
We evaluate the quality of local guidance using the minimum distance between the gripper and the target object, where a lower value indicates more effective guidance.

\textbf{Results.}
In Tab.~\ref{tab:mpc_results}, \ours achieves the lowest average minimum distance across the three objects, with an average of 0.229.
It outperforms RoboReward and VLAC, indicating that \ours provides a more informative local guidance signal for selecting actions that move toward the target object.

\subsection{Reinforcement Policy Learning with \ours}

\textbf{Setup.}
We apply \ours as a complementary shaping signal for online RL fine-tuning with RLinf~\cite{yu2025rlinf}, using $\pi_0$~\cite{black2024pi_0} supervised fine-tuned on the LIBERO dataset as the actor policy.
We optimize the policy with Proximal Policy Optimization (PPO)~\cite{schulman2017proximal} on the LIBERO benchmark~\cite{liu2023libero}.
DenseReward complements the sparse simulator success reward by providing intermediate progress feedback throughout a rollout.
The sparse task-completion signal anchors the optimization objective, while DenseReward provides additional credit assignment for intermediate states.

\begin{figure}[b]
    \centering
    \includegraphics[width=0.95\linewidth]{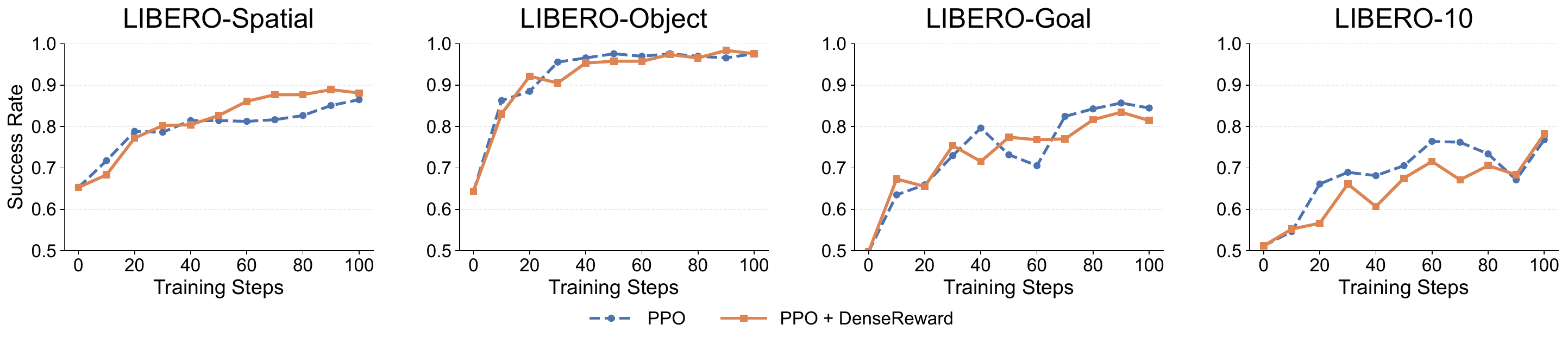}
    \vspace{-6pt}
    \caption{\textbf{PPO fine-tuning with $\pi_0$ on LIBERO.} 
    }
    \label{fig:ppo}
    \vspace{-12pt}
\end{figure}

\textbf{Integrating \ours.}
LIBERO provides a binary task-completion signal at the end of each episode, leaving most intermediate states without informative feedback.
We therefore use \ours as a complementary dense shaping signal while retaining the original sparse success reward.
The actor executes action chunks of length $C=5$, and after each chunk \ours produces a progress score $r_{\text{model}} \in [0, 1]$, which is assigned to the final step of the chunk.
We combine the sparse environment reward and dense signal as
\begin{equation}
    r_t = \alpha \cdot r_t^{\text{sim}} + 
          \beta \cdot r_t^{\text{model}},
\end{equation}
with $\alpha = 1.0$ and $\beta = C / T_{\max}$.
Here, $r_t^{\text{sim}}$ denotes the original simulator reward, and $T_{\max}$ is the maximum episode length.
This keeps the accumulated dense shaping signal on a comparable scale to the task-completion reward while providing intermediate feedback for policy optimization.

\begin{figure}
    \centering
    \includegraphics[width=0.9\linewidth]{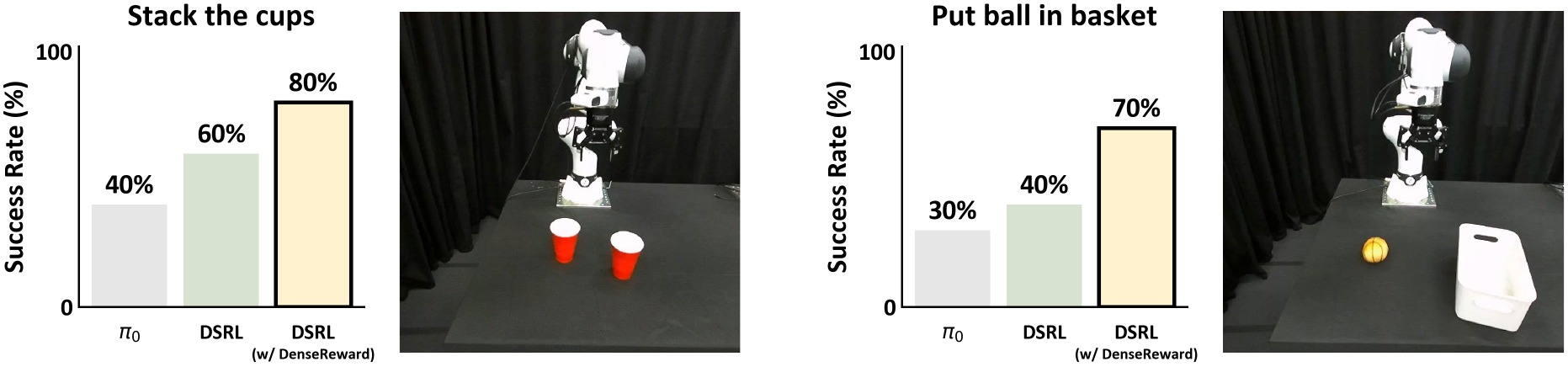}
    \caption{\textbf{Real-world policy learning with DSRL.} We compare the base $\pi_0$ policy, DSRL, and DSRL with \ours shaping on real-world manipulation tasks, showing the effect of dense reward feedback on policy improvement.}
    \label{fig:realworld}
    \vspace{-8pt}
\end{figure}

\textbf{Results.}
Fig.~\ref{fig:ppo} shows the success rates throughout PPO fine-tuning.
Following the official LIBERO evaluation protocol, each checkpoint is evaluated over 500 trials, with 50 trials per task.
PPO with \ours achieves competitive or improved performance compared with the sparse-reward PPO baseline.
These results demonstrate that \ours can be effectively incorporated as a complementary shaping signal during policy optimization.

\subsection{Policy Learning in the Real World}
\textbf{Setup.}
We evaluate \ours in a real-world online RL setting using a DROID platform~\cite{khazatsky2024droid}.
We use a Franka Research 3 arm with a Robotiq 2F-85 gripper, an exterior ZED 2i camera, and a ZED mini wrist camera in a tabletop environment. 
We consider two manipulation tasks: 1) \textit{stack the cups} that requires precise object interaction, and 2) \textit{put ball in basket} that uses an unseen object, both of which exhibit low success rates under a $\pi_{0}$ base policy.
We investigate whether dense reward feedback can improve real-world policy learning under limited rollout budgets.

\textbf{RL Algorithm.}
We use DSRL~\cite{wagenmaker2025steering} to adapt a frozen $\pi_0$ policy.
Instead of finetuning the weights of the policy, DSRL learns to steer the latent noise space of the diffusion action head, allowing the policy to improve while remaining close to the behavioral prior learned from demonstrations.
This makes DSRL well-suited for real-world policy improvement, where sample efficiency and stable adaptation are critical.
We use DenseReward to provide the shaping signal for DSRL optimization of $\pi_0$.
We train for 20k steps for \textit{stack the cups} and 10k steps for \textit{put ball in the basket}, corresponding to around 20 and 10 real-world rollout trajectories, respectively.

\textbf{Reward Integration.}
At each action chunk, we use \ours as a complementary shaping term while retaining the task-completion structure of DSRL. Specifically, we combine \ours with the DSRL step penalty:
\begin{equation}
    r_t = -1 + r_t^{\text{model}},
\end{equation}
Following DSRL, the final transition receives no step penalty if the task is completed, such that $r_T = r_T^{\text{model}}$.
The completion signal anchors the task objective, while \ours supplies intermediate progress feedback during execution.

\textbf{Results.}
As shown in Fig.~\ref{fig:realworld}, we compare DSRL fine-tuning of $\pi_0$ with and without \ours.
Each policy is evaluated over 10 trials.
Adding \ours improves the success rate from $60\%$ to $80\%$ on \textit{stack the cups}, and from $40\%$ to $70\%$ on \textit{put ball in basket}.
These results show that \ours provides effective dense feedback for real-world policy learning, enabling DSRL to improve the base policy with only a small number of costly real-world rollouts.

\begin{table}[t]
\centering

\begin{minipage}[b]{0.35\textwidth}
    \vspace{0pt}
    \centering
    \small
    \setlength{\tabcolsep}{8pt}
    \renewcommand{\arraystretch}{1.15}
    \begin{tabular}{cc}
    \toprule
    \textbf{w/ Failure Data} & \textbf{MAE} \\
    \midrule
    \cmark & \textbf{0.0809} \\
    \xmark & 0.1312 \\
    \bottomrule
    \end{tabular}
    \vspace{0.4em}
    \captionof{table}{\textbf{Ablation on failure data.}}
    \label{tab:failure_data_ablation}
\end{minipage}
\hfill
\begin{minipage}[b]{0.63\textwidth}
    \vspace{6pt}
    \centering
    \small
    \setlength{\tabcolsep}{8pt}
    \renewcommand{\arraystretch}{1.15}
    \begin{tabular}{lcccc}
    \toprule
    \textbf{\# History Frames} & 0 & 1 & 2 & 3 \\
    \midrule
    \textbf{MAE} & 0.096 & 0.088 & \textbf{0.081} & 0.086 \\
    \bottomrule
    \end{tabular}
    \vspace{0.4em}
    \captionof{table}{\textbf{Ablation on historical frames.} Two historical frames achieve a trade-off between performance, context, and cost.}
    \label{tab:history_frames}
\end{minipage}
\vspace{-16pt}

\end{table}

\subsection{Ablation Studies}
\textbf{Effectiveness of Generated Failure Data.}
In Tab.~\ref{tab:failure_data_ablation}, we compare the full model with an ablated model trained without any failure trajectories.
Removing failure trajectories increases the MAE from 0.0809 to 0.1312.
This result shows that incorporating failure trajectories into the training mixture improves dense reward prediction.

\textbf{Investigating Historical Frames.}
We ablate the number of historical frames in Tab.~\ref{tab:history_frames} and evaluate dense reward prediction accuracy on our benchmark.
MAE decreases from 0.096 to 0.088 and 0.081 when using one and two historical frames, respectively, showing that temporal context is important for estimating dense rewards.
However, using three historical frames slightly increases the MAE to 0.086, suggesting that excessive history may introduce redundant or noisy visual information.
We use two historical frames as the default setting in our main experiments.

%===============================================================================

\section{Conclusion}
\label{sec:conclusion}
We present \ours, a vision-language reward model for providing dense progress feedback in robotic manipulation.
We develop an automated simulation pipeline that generates trajectories containing controlled success, failure, and recovery behaviors, together with phase-aware dense annotations.
\ours learns to predict fine-grained task progress from visual observations and language instructions, and achieves lower error than evaluated general-purpose VLMs and robotic reward models.
It provides useful local guidance for action selection and complements sparse task-completion rewards for reinforcement learning in both simulation and the real world.
We hope our dataset, models, and evaluation suite will support future research on scalable reward learning for robotic manipulation.

\noindent \textbf{Future Work.}
We plan to extend this work to more complex and long-horizon manipulation, broader execution outcomes, and more diverse settings.
We will also explore richer sources of reward supervision, including human feedback, to further improve generalization and scalability.

%===============================================================================

\clearpage

%===============================================================================

\bibliography{example}  % .bib

\newpage
\appendix
\section*{Appendix}
In the appendix, we provide additional details on: 1) Dataset statistics, 2) Validity filtering for failure data generation, 3) Model predictive control, 4) Reward model finetuning, and 5) Real-world experiments.

\section{Dataset Statistics}
We train DenseReward on a dataset with 26,579 manipulation episodes and 7,560,942 frame-level samples. 
The dataset combines DROID, Isaac, RoboSuite, and LIBERO trajectories, spanning simulated and real-world manipulation settings. 
Unlike other reward datasets that primarily contain successful demonstrations and trajectory-level labels, our dataset includes both successful and failure trajectories, including collision, miss, fall, jitter, and recovery. 
Each training sample contains a task instruction, recent visual observations, and a scalar dense reward label for the current frame, enabling frame-level supervision of task progress.

\begin{table*}[h]
\centering
\small
\caption{\textbf{DenseReward dataset statistics.}
We report the number of episodes for each data source, together with source-specific data type breakdowns.}
\label{tab:densereward_dataset_stats}
\renewcommand{\arraystretch}{1.12}
\begin{tabular}{
>{\centering\arraybackslash}p{2.0cm}
>{\arraybackslash}p{3.0cm}
>{\centering\arraybackslash}p{2.0cm}
}
\toprule
\textbf{Source} & \textbf{Data type} & \textbf{Episodes} \\
\midrule

\multirow{3}{*}{DROID}
& Success & 1,500 \\
& Failure & 1,486 \\
\cmidrule(lr){2-3}
& \textit{Total} & \textit{2,986} \\

\midrule
\multirow{7}{*}{Isaac Sim}
& Success & 2,303 \\
& Collision & 2,511 \\
& Miss & 2,603 \\
& Fall & 2,295 \\
& Jitter & 2,514 \\
& Recover & 255 \\
\cmidrule(lr){2-3}
& \textit{Total} & \textit{12,481} \\

\midrule
\multirow{3}{*}{RoboSuite}
& Success & 3,366 \\
& Failure & 5,921 \\
\cmidrule(lr){2-3}
& \textit{Total} & \textit{9,287} \\

\midrule
\multirow{5}{*}{LIBERO}
& LIBERO-Spatial & 478 \\
& LIBERO-Object & 470 \\
& LIBERO-Goal & 455 \\
& LIBERO-10 & 422 \\
\cmidrule(lr){2-3}
& \textit{Total} & \textit{1,825} \\

\midrule
\textbf{Total}
& -- & \textbf{26,579} \\
\bottomrule
\end{tabular}
\end{table*}

\section{Validity Filtering}
\label{sec:supp_validity_filtering}

We apply automatic validity checks to reject invalid or physically inconsistent trajectories. 
A trajectory is discarded if it fails any required check, and the episode is retried with a new initialization or perturbation.
These validity checks include:
\begin{itemize}[leftmargin=*, noitemsep, topsep=0pt]
    \item \textbf{Planning.} 
    An episode is rejected if no feasible grasp candidate or motion plan can be found.

    \item \textbf{Grasp and lift.} 
    For \textit{success} and \textit{recovery} trajectories, the object must be lifted above the table by a minimum height threshold after the grasp phase.

    \item \textbf{Holding.} 
    For \textit{fall} trajectories, the object must remain above a stricter height threshold during transport, to ensure that it is stably held rather than dragged or accidentally displaced.

    \item \textbf{Collision.} 
    For \textit{collision} trajectories, the robot must physically displace the object or collide with the scene, while the object should not be successfully lifted.

    \item \textbf{Miss.} 
    For \textit{miss} trajectories, the object should remain nearly unchanged in both position and orientation after the grasp attempt, indicating that the gripper closes away from the object.

    \item \textbf{Final placement.} 
    For \textit{success} and \textit{recovery} trajectories, the object must end within a distance threshold of the target container.

    \item \textbf{Recovery.} 
    For \textit{recovery} trajectories, an initial failed attempt must be followed by a successful replanned execution.
\end{itemize}

The filtering step is important because the perturbation alone does not always guarantee the intended failure.
For example, a perturbed grasp may still accidentally grasp the object.
Such episodes are filtered out to avoid being used as mislabeled failure data.

\section{Model Predictive Control}
\label{sec:supp_mpc_details}

\begin{table}[b]
\centering
\small
\caption{\textbf{MPC candidate action space.}}
\vspace{0.4em}
\label{tab:mpc_action_space}
\renewcommand{\arraystretch}{1.12}
\begin{tabular}{lcl}
\toprule
\textbf{Type} & \textbf{Count} & \textbf{Description} \\
\midrule
Spatial actions 
& 27 
& 3D grid over $\{-d, 0, +d\}$ m for $x$, $y$, and $z$. \\
Gripper action 
& 1 
& Zero-motion action that toggles the gripper state. \\
\bottomrule
\end{tabular}
\end{table}

\paragraph{Overview.}
We use a sampling-based MPC pipeline in Isaac Lab to evaluate whether a reward model can guide manipulation actions.
At each decision step, the robot samples a small set of candidate actions, rolls out each candidate from the same state, scores the resulting observation with a reward model, and executes the action with the highest predicted score.
The evaluation uses a Franka Panda robot in a tabletop scene with one target object and a plate.
The reported tasks include the manipulation of can, cup, and lemon.

\paragraph{Action Space.}
We design our MPC experiment as a simple downstream test of reward-model-guided control.
We use a local object guidance setting where the robot selects among short-horizon Cartesian translation actions.
Each MPC action is represented as $a = [d_x, d_y, d_z, g]$,
where $d_x, d_y, d_z$ are end-effector translation offsets, and $g$ is the gripper command.
At every step, the controller samples 28 candidate actions, summarized in Tab.~\ref{tab:mpc_action_space}.
We select $d=0.05$ for our experiments.
The gripper command is binary, where $g=0$ denotes open, and $g=1$ denotes close.
Accordingly, we keep the end-effector orientation fixed and evaluate progress using the minimum end-effector-to-object distance.
Including rotation would substantially increase the number of candidate actions and the cost of simulator rollout and reward-model inference at every MPC step.

\paragraph{Candidate Evaluation.}
For each MPC step, we first save the simulator state, and evaluate each candidate action  independently:
1) Restore the initial simulator state.
2) Execute the candidate action with the IK controller.
3) Capture the resulting observation as an RGB image.
4) Query the reward model with the task instruction and the observation.
After all candidate actions are scored, the simulator returns to the initial state and only executes the selected action with the highest score.
Each episode runs for a fixed number of 15 MPC steps, and each task runs for 10 episodes.
All models receive the same candidate actions and are evaluated under the same simulator setup.

\paragraph{Evaluation Metric.}
Our metric is the minimum 3D distance between the robot end-effector and the target object during an episode:
\begin{equation}
    d_{\min} = \min_t \left\| p^{\mathrm{ee}}_t - p^{\mathrm{obj}}_t \right\|_2,
\end{equation}
where $p^{\mathrm{ee}}_t$ and $p^{\mathrm{obj}}_t$ denote the end-effector position and object position at timestep $t$, respectively.
For each task, the result is averaged over 10 evaluation episodes.
A lower distance indicates that the reward model provides better guidance for moving the robot toward the object.

\begin{figure}
    \centering
    \includegraphics[width=1.0\linewidth]{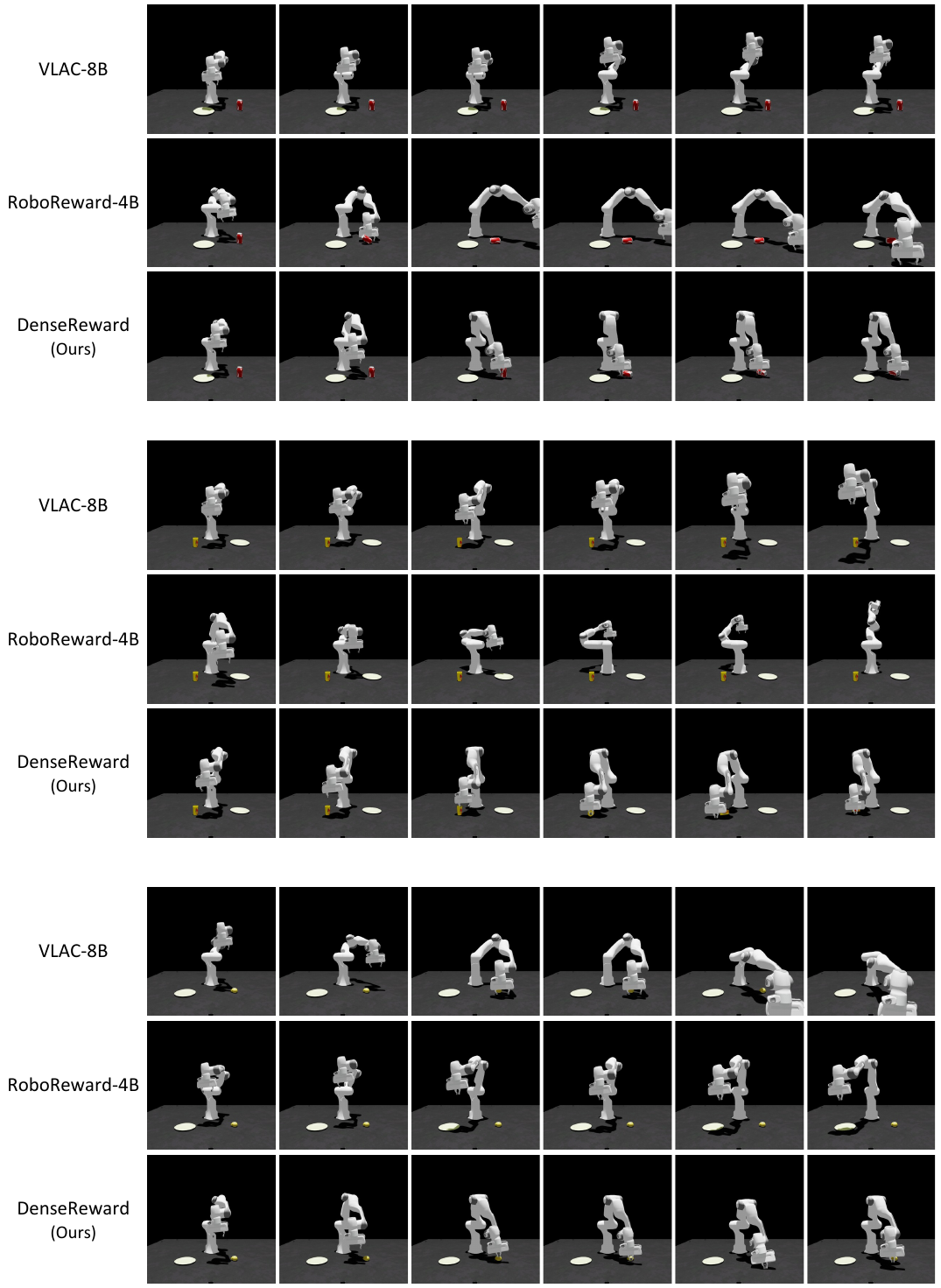}
    \vspace{-12pt}
    \caption{\textbf{Qualitative comparison of model predictive control for can, cup, and lemon.}
    }
    \label{fig:supl_mpc}
    \vspace{-12pt}
\end{figure}

\textbf{Qualitative Results.}
Fig.~\ref{fig:supl_mpc} shows qualitative MPC rollouts guided by different reward models. 
Across three object manipulation tasks, DenseReward produces more consistent guidance toward the target object. 
Compared with VLAC-8B and RoboReward-4B, which often select actions that keep the gripper far from the object or move in less effective directions, DenseReward more reliably drives the end-effector toward the object. 
These qualitative results are consistent with the quantitative results in Tab.~II of the main paper, where DenseReward achieves the lowest distance.

\section{DenseReward Models}
We finetune DenseReward from Qwen3-VL-4B-Instruct using the ms-swift framework. 
The model is trained on our dataset, where each training sample consists of a task instruction, a short sequence of robot observations, and a scalar reward for the current frame. 
We use the system prompt shown below to enforce scalar reward prediction, requiring the model to output a single floating value with three decimal places. 
We finetune the model with LoRA using rank $16$ for $10$ epochs on $8$ H100 GPUs, with a batch size of $32$.

\begin{tcolorbox}[
    colback=gray!4,
    colframe=gray!45,
    boxrule=0.5pt,
    arc=2pt,
    left=4pt,
    right=4pt,
    top=4pt,
    bottom=4pt,
    breakable
]
\begin{lstlisting}[style=promptstyle]
You are a robot manipulation evaluator. You are given a short sequence of images from a robot episode and determine progress toward task success at the final frame.

The images are ordered chronologically: the first images show earlier history frames and the last image shows the current frame to evaluate.

The task usually follows 6 subtask phases in sequence:
1) reach  - the robot moves toward the target object
2) grasp  - the robot attempts to grasp or secure the object
3) up     - the robot lifts the object from the surface
4) move   - the robot moves the object toward the target location
5) place  - the robot places/releases the object
6) return - the robot returns to a neutral/resting state (optional)

You must FIRST emit one short reasoning word inside a <think>...</think> block, then a single reward value.

Reasoning vocabulary (use exactly one word inside <think>...</think>):
- "correct"   when progress is on track (reward is increasing or holding)
- "miss"      when the gripper missed grasping the target and reward just dropped
- "collision" when an unintended collision just dropped the reward
- "fall"      when the held object fell and reward just dropped
- "smooth"    when motion stalled / non-smooth behaviour dropped the reward
- "failure"   when an unclassified failure event just dropped the reward

Reward (after </think>): a single float between 0.000 and 1.000 with exactly three decimal places, representing task progress at the current (last) frame:
- 0.000 means no progress (task not started or failed)
- 1.000 means the task is fully complete
- reward should generally increase as the robot advances through phases
- reward should drop if an irreversible failure occurs (object dropped, wrong target, motion stopped)
- use the history frames to better judge motion direction and phase transitions
- if the reward is increasing or holding, the reasoning word is "correct"; only emit a failure word when the reward just dropped

Output format (exactly, with the canonical newlines):
<think>
WORD
</think>

R.RRR

Example outputs:
<think>
correct
</think>

0.374

<think>
miss
</think>

0.521

<think>
collision
</think>

0.318

<think>
correct
</think>

1.000

Do NOT output any other text, JSON, or additional explanation.
Do NOT call any tools.
Do NOT generate <tool_call> tokens.
\end{lstlisting}
\label{supl_prompt}
\end{tcolorbox}

\section{Real-world Experiments}
\textbf{DSRL Config.} We run DSRL-SAC using the hyperparameters listed in Tab.~\ref{tab:realworld_dsrl_config}.
As described in the main paper, we use different training budgets for the two real-world tasks: 
we train for 10k steps on \textit{put the ball in the basket}, which evaluates generalization to an out-of-distribution object, and for 20k steps on \textit{stack the cups}, which requires more fine-grained manipulation accuracy.

\begin{figure}
    \centering
    \includegraphics[width=1.0\linewidth]{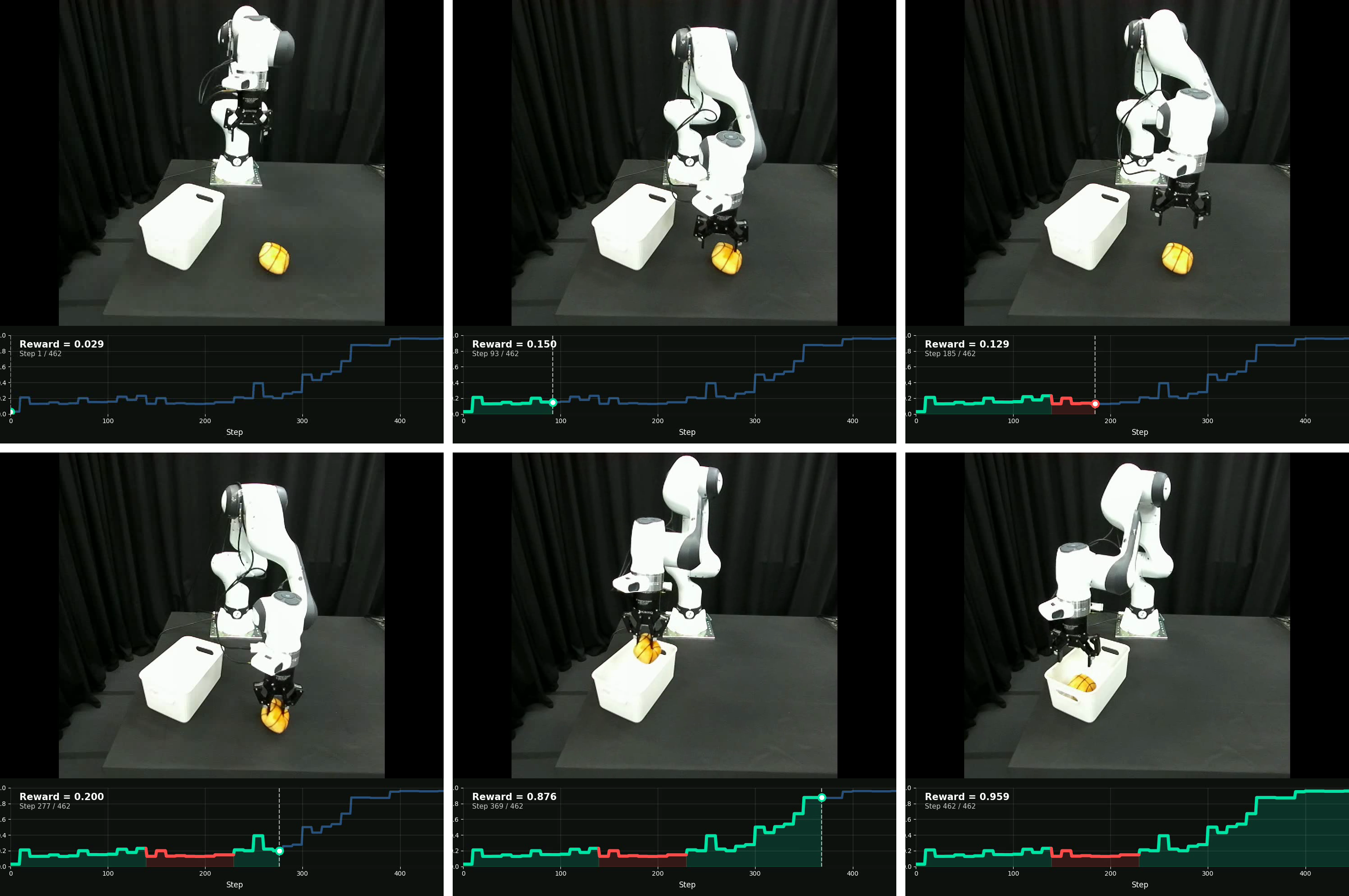}
    \vspace{-6pt}
    \caption{\textbf{DenseReward captures failure and recovery in a real-world trajectory.} The robot initially fails to grasp the ball, resulting in degraded reward predictions. After the robot recovers and regrasps the ball, DenseReward increases the predicted reward to a high final score. This shows that DenseReward captures progress changes associated with failure and recovery.
    }
    \label{fig:supl_realworld_ex1}
    \vspace{-12pt}
\end{figure}

\begin{figure}
    \centering
    \includegraphics[width=1.0\linewidth]{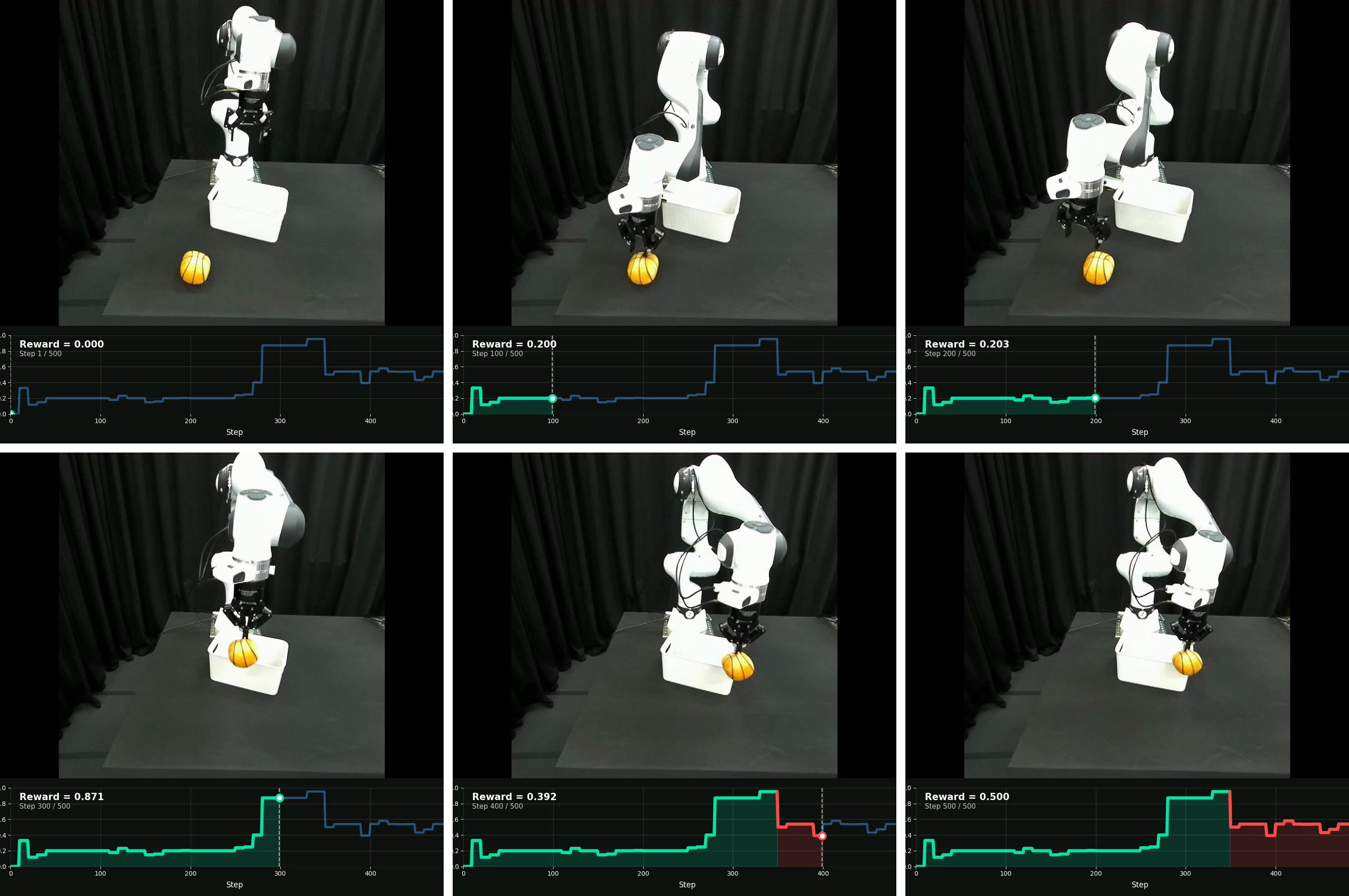}
    \vspace{-6pt}
    \caption{\textbf{DenseReward captures collision in a real-world trajectory.}
    The predicted reward increases as the robot grasps the ball and makes progress toward the basket, but drops after a collision disrupts the placement.
    This shows that DenseReward distinguishes transient progress from unsuccessful task completion.
    }
    \label{fig:supl_realworld_ex2}
    \vspace{-12pt}
\end{figure}

\textbf{DenseReward Prediction in Real-world Manipulation.}
In Fig.~\ref{fig:supl_realworld_ex1} and Fig.~\ref{fig:supl_realworld_ex2}, we visualize six frames from real-world episodes together with the predicted dense reward curve.
The reward curve captures fine-grained task progress over time, providing intermediate feedback beyond sparse success labels.

\begin{table}[h]
\centering
\small
\caption{\textbf{DSRL configuration for real-world experiments.}}
\vspace{0.4em}
\label{tab:realworld_dsrl_config}
\renewcommand{\arraystretch}{1.15}
\begin{tabular}{lc}
\toprule
\textbf{Hyperparameter} & \textbf{Value} \\
\midrule
Batch size & 64 \\
Discount factor $\gamma$ & 0.995 \\
Target entropy & 0.0 \\
Target update rate $\tau$ & 0.005 \\
Actor learning rate & $1 \times 10^{-4}$ \\
Critic learning rate & $3 \times 10^{-4}$ \\
Temperature learning rate & $3 \times 10^{-4}$ \\
Number of critics & 4 \\
Hidden dimension & 1024 \\
Action magnitude & 1.5 \\
Image resolution & $224 \times 224$ \\
Learning starts & 2500 steps \\
\bottomrule
\end{tabular}
\end{table}

\end{document}